\documentclass[10pt,twocolumn,letterpaper]{article}

\usepackage{iccv}
\usepackage{times}
\usepackage{epsfig}
\usepackage{graphicx}
\usepackage{amsmath}
\usepackage{amssymb}
\usepackage{algorithm,algorithmic,setspace}
\usepackage{fmtcount}

\usepackage[pagebackref=true,breaklinks=true,letterpaper=true,colorlinks,bookmarks=false]{hyperref}

 \iccvfinalcopy 

\def\iccvPaperID{1193} 

\def\BE{\vspace{-2.0mm}\begin{equation}}
\def\EE{\vspace{-1.4mm}\end{equation}}

\def\BEA{\vspace{-2.0mm}\begin{eqnarray}}
\def\EEA{\vspace{-1.0mm}\end{eqnarray}}

\newcommand{\conv}{*}

\newcommand{\eqn}[1]{Eqn.~\ref{eqn:#1}}
\newcommand{\fig}[1]{Fig.~\ref{fig:#1}}
\newcommand{\tab}[1]{Table~\ref{tab:#1}}
\newcommand{\secc}[1]{Section~\ref{sec:#1}}
\def\etal{{\textit{et~al.~}}}

\ificcvfinal\fi
\begin{document}

\title{Differentiable Pooling for Hierarchical Feature Learning}

\author{Matthew D.\ Zeiler and Rob Fergus\\
Dept.~of Computer Science, Courant Institute, New York University\\
{\tt\small \{zeiler,fergus\}@cs.nyu.edu}}

\maketitle
\thispagestyle{empty}

\begin{abstract}
  We introduce a parametric form of pooling, based on a Gaussian,
  which can be optimized alongside the
  features in a single global objective function. By contrast,
  existing pooling schemes are based on heuristics (e.g.~local
  maximum) and have no clear link to the cost function of the
  model. Furthermore, the variables of the Gaussian explicitly store
  location information, distinct from the appearance captured by the
  features, thus providing a what/where decomposition of the input
  signal. Although the differentiable pooling scheme can be
  incorporated in a wide range of hierarchical models, we demonstrate
  it in the context of a Deconvolutional Network model (Zeiler \etal
  \cite{Zeiler11}). We also explore a number of secondary issues
  within this model and present detailed experiments on MNIST digits. 
\end{abstract}

\vspace{-4mm}
\section{Introduction}

A number of recent approaches in vision and machine learning have
explored hierarchical representations for images and video, with the
goal of learning features for object recognition. One class of
methods, for example Convolutional Neural Networks \cite{Lecun1989} or the recent
RICA model of Le \etal \cite{Le12}, use a purely feed-forward
hierarchy that maps the input image to a set of features which are
presented to a simple classifier. Another class of models attempts to
build hierarchical generative models of the data. These include Deep
Belief Networks \cite{Hinton2006b}, Deep Boltzmann Machines \cite{SalHinton07}
and the Compositional Models of Zhu \etal \cite{Zhu10,Chen07}.


Spatial pooling is a key mechanism in all these hierarchical image
representations, giving invariance to local perturbations of the
input and allowing higher-level features to model large portions of
the image. Sum and max pooling are the most common forms, with max
being typically preferred (see Boureau \etal \cite{ylan10} for an
analysis).

In this paper we introduce a parametric form of pooling that can be
directly integrated into the overall objective function of many
hierarchical models. Using a Gaussian parametric model, we can
directly optimize the mean and variance of each Gaussian pooling region
during inference to minimize a global objective function. This contrasts with existing
pooling methods that just optimize a local criterion (e.g.~max over a
region). Adjusting the variance of each Gaussian allows a smooth
transition between selecting a single element (akin to max
pooling) over the pooling region, or averaging over it (like a sum
operation). 

Integrating pooling into the objective facilitates joint training and
inference across all layers of the hierarchy, something that is often
a major issue in many deep models. During training, most approaches
build up layer-by-layer, holding the output of the layer beneath
fixed. However, this is sub-optimal, since the features in the
low-layers cannot use top-down information from a higher layer to
improve them. A few approaches do perform full joint training of the
layers, notably the Deep Boltzmann Machine \cite{SalHinton07},
and Eslami \etal \cite{Eslami12}, as applied to images, and the Deep
Energy Models of Ngiam \etal \cite{Ngiam11}.  We demonstrate our
differentiable pooling in a third model with this capability, the
Deconvolutional Networks of Zeiler \etal \cite{Zeiler11}. This is a
simple sparse-coding model that can be easily stacked and we show how
joint inference and training of all layers is possible, using the
differentiable pooling. However, differentiable pooling is not
confined to the Deconvolutional Network model -- it is capable of
being incorporated into many existing hierarchical models.

The latent variables that control the Gaussians in our pooling scheme
store location information (``where''), distinct from the features that
capture appearance (``what''). This separation of what/where is also
present in Ranzato \etal \cite{Ranzato07}, the transforming
auto-encoders of Hinton \etal \cite{Hinton11}, and Zeiler \etal
\cite{Zeiler11}.

In this paper, we also explore a number of secondary issues that help
with training deep models: non-negativity constraints; different forms
of sparsity; overcoming local minima during inference and different
sparsity levels during training and testing. 





\begin{figure*}[t!]
\begin{center}
\includegraphics[width=7in]{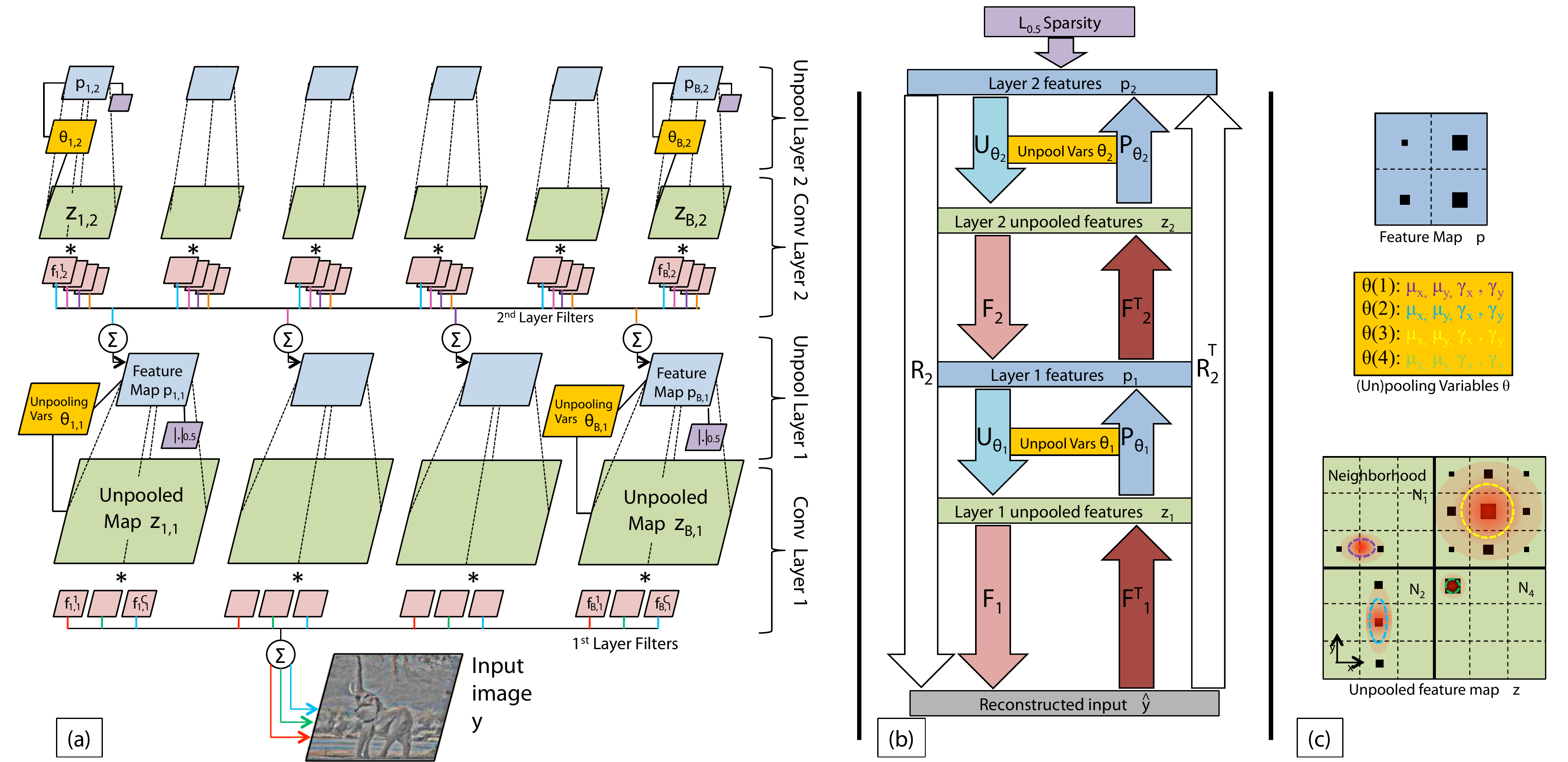}
\end{center}
\vspace{-4mm}
\caption{(a): A 2-layer model architecture.  (b): Schematic of inference in a two layer model. (c): Illustration of the Gaussian parameterization
  used in our differentiable pooling.}
\label{fig:big} 
\vspace{-4mm}
\end{figure*}


\section{Model Overview}
We explain our contributions in the context of a
Deconvolutional Network, introduced by Zeiler \etal
\cite{Zeiler11}. This model is a hierarchical form of convolutional sparse coding
that can learn invariant image features in an unsupervised
manner. Its simplicity allows the easy integration of differentiable
pooling and is amenable to joint inference over all layers.

Let us start by reviewing a single Deconvolutional Network layer, presented
with an input image $v^c$ (having $c$ color channels). The goal
is to produce a reconstruction $\hat{v}$ from sparse features $p$,
that is close to $v$. We achieve this by minimizing: 
\begin{equation}
\frac{\lambda}{2} \sum_c  \| \hat{v}^c - v^c \|^2_2 + |p|_\alpha
\label{eqn:sl}
\end{equation} 
where $\lambda$ is a hyper-parameter that controls the influence of
the reconstruction term. $p$ consists of a set of $B$ 2-D feature maps, thus forming an over
complete-basis. To give a unique
solution, a sparsity constraint on $p$ is needed and we use an element-wise pseudo-norm
where $0.5 \leq \alpha \leq 1$. The reconstruction $\hat{v}$ is produced
from $p$ by two sub-layers: {\em Unpooling} and {\em Convolution}.

\subsection{Unpooling}
\label{sec:unpool}
In the unpooling sub-stage, each 2D feature map $p_b$ undergoes an
unpooling operation to produce a larger 2D {\em unpooled} feature map
$z_b$\footnote{3D (un)pooling is also possible, as explored in
  \cite{Zeiler11}.}. Each element $j$ in $p_b$ influences a small
neighborhood $N_j$ (typically $2 \times 2$ or $3 \times 3$) in the
unpooled map $z_b$, via a set of weights $w(i)$ within the
neighborhood: 
\BE z_b(i) = w(i) p_k(j) \quad \forall i \in N_j
\label{eqn:unpool}
\EE
We constrain the weights $w(i)$ to have unit $\ell_2$-norm, as this
makes the unpooling operation invertible\footnote{Combining
  Eqs.~\ref{eqn:unpool} and \ref{eqn:pool}, we have $p_b(j) = \sum_i
  w^2(i) p_b(j)$, hence $\sum_i
  w^2(i)$=1.}. The inverse {\em pooling}
operation computes each element $j$ in $p_b$ as the sum of weights
$w(i)$ in neighborhood $N_j$ of the unpooled map $z_b$:
\BE
p_b(j) = \sum_{i \in N_j} w(i) z_b(j) 
\label{eqn:pool}
\EE
In Zeiler \etal \cite{Zeiler11}, max (un)pooling was used,
equivalent to $w(i)$ being all zero, except for a single element set
to 1. In this work, we consider more general $w(i)$'s, as detailed in
\secc{diff_pool}, treating them as latent variables which will be
inferred for each input image. Note that each element in $p$ has its
own set of $w$'s.

For the rest of the paper, we consider the neighborhoods $N_j$ to be
non-overlapping, but the above formulation generalizes to overlapping
regions as well. For brevity, we write the unpooling operation as a single linear
matrix, parameterized by weights $w$: $z = U_w p$.

\subsection{Convolution}
In the convolution sub-stage, the reconstruction $\hat{v}^c$ is
formed by convolving 2D unpooled feature maps $z_b$ with filters
$f^c_b$ and summing them:
\begin{equation}
\hat{v}^c = \sum_{b=1}^B z_b \conv f^c_b 
\label{eqn:deconv}
\end{equation} 
where $\conv$ is the 2D convolution operator. The filters $f$ are the
parameters of the model common to all images. The feature maps $z$ are
latent variables, specific to each image.  For notational brevity, we
combine the convolution and summing operations into a single
convolution matrix $F$ and convert the multiple 2D maps $z_{b}$ into a
single vector $z$: $\hat{v} = F z$.

\subsection{Discussion of Single Layer}
The combination of unpooling and convolution operations gives the
reconstruction $\hat{v}$ from $p$:
\BE
\frac{\lambda}{2} \| F U_w p - v \|_2^2 + | p |_\alpha
\label{eqn:single}
\EE 
A single layer of the model is shown in the lower part of \fig{big}(a). This
integrated formulation allows the straightforward optimization of the
filters $f$, features $p$ and the (un)pooling weights $w$ to minimize
a single objective function. While most other models also learn
filters and features, the pooling operation is typically fixed. Direct
optimization of \eqn{single} with respect to $w$ is one the main
contributions of this work and is described in \secc{diff_pool}.

Note that, given fixed weights $w$, the reconstruction is linear in
$p$, thus \eqn{single} describes a tri-linear model, with $w$
coding position (where) information about the (what) features $p$. 

\eqn{single} differs from the original Deconvolutional Network
formulation \cite{Zeiler11} in several important ways. First, sparsity is imposed directly
on $p$, as opposed to $z$. This integrates pooling into the objective
function, allowing it to become part of the inference. Second,
\cite{Zeiler11} considers only $\alpha=1$, rather than the
hyper-Laplacian ($\alpha<1$) sparsity we employ. Third, $p$ is
non-negative, as opposed to \cite{Zeiler11}  where there was no such
constraint. Fourth, and most importantly, by inferring the optimal
(un)pooling weights $w$ we directly minimize the objective function of
the model. Fixed sum or max pooling, employed by other approaches, is
a local heuristic that has no clear relationship to the overall cost.

\subsection{Multiple Layers}
Multi-layer models are constructed by stacking the single layer model
described above in the same manner as Zeiler \etal
\cite{Zeiler11}. The feature maps $p$ from one layer become the input
maps to the layer above (which now has $B$ ``color channels'').

An important property of the model is that feature maps exist solely
at the top of the model (there are no explicit features in intermediate
layers), thus the only variables at the intermediate layers are
filters $F$ and unpooling weights $w$.
For an $l$ layer model, the reconstruction $\hat{v}$ is:
\BE
\hat{v} = F_1 U_{w_1} F_2 U_{w_2} \ldots F_l U_{w_l} p_l = R_l p_l 
\EE 
 where $F_k$ and $U_{w_k}$ are the convolutional and unpooling
 operations from each layer $k$. We condense the sequence of unpooling and
 convolution operations into a single {\em reconstruction} operator
 $R_l$, which lets us write the overall object for a multi-layer model (shown here for a single image, $v$, but optimized over a set of images $v^1,\ldots,v^I$ during training):
\BE
\label{eqn:multi}
C_l(v) = \frac{\lambda}{2} \|R_l p_l - v \|_2^2 + |p_l|_{\alpha}
\EE
A multi-layer model is shown in \fig{big}(a). 
Note that since $R_l$ is linear, given the (un)pooling weights $w$,
the reconstruction term is easily differentiable. The derivative of
$R_l$ is simply $R^T_l = U^T_{w_l} F^T_l  \ldots U^T_{w_1} F^T_1$, which is a {\em forward propagation}
operator. This takes a signal at the input and repeatedly convolves
(using with flipped versions of the filters at each layer) and pools
(using weights $w_l$) all the way up to the features. This is a key
operation for both inference and learning, as described in
\secc{inference} and \secc{learning} respectively. \fig{big}(b)
illustrates the reconstruction and forward propagation operations.
 

 


\subsection{Differentiable Pooling} 
\label{sec:diff_pool}
We impose a parametric form on the (un)pooling weights $w$ to ensure
that the features are invariant to small changes in the input. The
pooling would otherwise be able to memorize perfectly the unpooled
features, giving ``lossless'' pooling which would not generalize at all. 

The parametric model we use is a 2D axis-aligned Gaussian, with mean
$(\mu_x,\mu_y)$ and precision $(\gamma_x,\gamma_y)$ over the pooling neighborhood $N_j$, introduced in \secc{unpool}. 
The Gaussian is normalized within the extent of the pooling region to
give weights $w$ whose square sums to 1 (thus giving unit $\ell_2$ norm):
\BE
\label{eqn:normGauss}
w(i) = \frac{\sqrt{a(i)}}{\sqrt{\sum_{i'} a(i')}}
\EE
where $a(i)$ is value of the Gaussian for element $i$, at location
$x(i), y(i)$ within the neighborhood $N_j$:
\BE
\label{eqn:gauss}
a(i) = e^{- [\frac{\gamma_x}{2} (x(i)-\mu_x)^2 + \frac{\gamma_y}{2}
  (y(i)-\mu_y)^2]} 
\EE 
\fig{big}(c) shows an illustration of this parameterization. For brevity, we let $\theta_j =
\{\mu_x,\mu_y,\gamma_x,\gamma_y\}$ be the parameters for neighborhood
$N_j$. We thus rewrite the unpooling operation in $U_{w_l}$ as
$U_{\theta_l}$. The Gaussian representation has several advantages
over existing sum or max pooling:
\begin{itemize}
\item Varying the mean of the Gaussian selects a particular region in
  the unpooled feature map, just like max pooling. This makes the
  feature invariant to small translations within the unpooled maps. 
\item Varying the precision of the Gaussian allows a smooth variation
  between max and sum operations (high and low precision
  respectively).
\item Changes in precision allow invariance to small scale
  changes in the unpooled features. For example, the width of an edge
  can easily be altered by adjusting the variance (see \fig{composite}(c)). 
\item The continuous nature of the Gaussian allows sub-pixel
  reconstruction that avoids aliasing artifacts, which can occur with
  max pooling. See \fig{maxgauss} for an illustration of this. 
\item The Gaussian representation is differentiable, i.e.~the
  gradient of \eqn{single} with respect to $\theta_j$ has analytic
  form, as detailed in \secc{w_infer}. 
\end{itemize}




\subsection{Non-Negativity} 
In standard sparse coding and other learning methods both the feature
activations and the learned parameters can be positive or
negative. This contrasts with our model, in which we enforce
non-negativity.

This is motivated by several factors. First, there is no notion of a
negative intensities or objects in the visual world. Second, the
Gaussian parameterization used in the differentiable pooling scheme,
described in \secc{diff_pool} has positive weights, so cannot
represent individual negative values in the unpooled feature maps. Third, there is
some biological evidence for non-negative representations within the
brain \cite{Hoyer2003}. Finally, we find experimentally that
non-negativity reduces the flexibility of the model, encouraging it
to learn good representations. The features computed at test-time have improved
classification performance, compared with models without this
constraint (see \secc{pos}).



\subsection{Hyper-Laplacian Sparsity} 
Most sparse coding models utilize the $\ell_{1}$-norm to enforce a
sparsity constraint on the features \cite{Olshausen1997}, as a proxy
for optimizing $\ell_{0}$ sparsity \cite{Tibshirani96}. However, a drawback of this form
of regularization is that it gives the same cost to two elements being 0.5
versus a single elements at 1 and the other at 0, even though the
latter has a lower $\ell_{0}$ cost. 

To encourage features with lower $\ell_{0}$ cost, we use a pseudo-norm
$\ell_{0.5}$ (i.e.~$\alpha=0.5$ in \eqn{single}) inspired by Krishnan and Fergus \cite{Krishnan09}, which aggressively pushes small elements toward zero. To optimize this, we experimented with techniques in \cite{Krishnan09}, but settled on gradient descent for simplicity.


\section{Inference}
\label{sec:inference}
\vspace{-2mm}
During inference, the filters $f$ at all layers are fixed and the
objective is to find the features $p$ and (un)pooling variables $\theta$
for all neighborhoods and all layers that minimize \eqn{multi}. We do this by
alternating between updating the features $p$ and the Gaussian
variables $\theta$, while holding the other fixed. 

\subsection{Feature Updates} 
\label{sec:p_infer}
For a given layer $l$, we seek the features $p_l$ that minimize
$C_l(v)$ (\eqn{multi}), given an input image $v$, filters
$f_1,\ldots,f_l$ and unpooling variables
$\theta_1,\ldots,\theta_l$. This is a large convolutional sparse
coding problem and we adapt the ISTA scheme of Beck and Teboulle
\cite{Beck09}. This uses an iterative framework of gradient and shrinkage steps. 

{\bf Gradient step:} The gradient of $C_l(v)$ with respect to $p_l$
is: 
\BE
\nabla p_l = \frac{\partial C_l(v)}{\partial p_l} = R^T_l (R_l p_l - v) 
\EE 
This involves first reconstructing the input from the current features: $\hat{v}=R_l p_l$,
computing the error signal $e=(\hat{v} - v)$, and then forward
propagating this up to compute the top layer gradient $\nabla p_l = R^T_l
e$. Given the gradient, we then can update $p_l$: 
\BE p_l = p_l -  \lambda_l \beta_{p_l} \nabla p_l
\label{eqn:grad}
\EE 
where the $\beta_{p_l}$ parameter sets the size of the gradient step. 

{\bf Shrinkage step:} Following the gradient step, we perform a per-element
shrinkage operation that clamps small elements in $p_l$ to zero, increasing its
sparsity. For $\alpha=1$, we use the standard $\ell_1$ shrinkage: 
\BE
p_l = \max(|p_l|-\beta_{p_l},0) \cdot \text{sign}(p_l)
\label{eqn:shrink}
\EE
For $\alpha=0.5$, we step in the direction of the gradient:
\BE
p_l = p_l-\beta_{p_l}  \frac{1}{2} \sqrt{|p_l|}^{-1}\cdot \text{sign}(p_l)
\label{eqn:shrinkL05}
\EE

{\bf Projection step:}
After shrinking small elements away, the solution is then projected onto the non-negative set:
\BE
p_l = \max(p_l,0)
\label{eqn:proj}
\EE

{\bf Step size calculation:} In order to set a learning rate for the
feature map optimization, we employ an estimation technique for
steepest descent problems \cite{Shewchuk1994} which uses the
gradients $\nabla p_l = \frac{\partial C}{\partial p_l}$ : 
\BE \beta_{p_l} = \frac{\nabla p_l^T \nabla p_l}{\nabla p_l^T R_l^T R_l \nabla p_l }
\label{eqn:autostep}
\EE
Automating the step-size computation has two advantages. First, each
layer requires a significantly different learning rate on account of
the differences in architecture, making it hard to set
manually. Second, by computing the step-size 
before each gradient step, each ISTA iteration makes good progress at
reducing the overall cost. In practice, we find fixed step-sizes to be
significantly inferior.

 $\nabla p_l$ is computed once per mini-batch. For
efficiency, instead of computing the denominator in \eqn{autostep} for each image,
we estimate it by selecting a small portion ($\sim$10\%) of each mini-batch.

{\bf Reset step:} Repeated optimization of the objective function
tends to get stuck in local minima as it proceeds over the dataset for
several epochs. We found a simple and effective way to overcome this
problem. By setting all feature maps $p_l$ to 0 every few epochs
(essentially re-initializing inference), cleaner filters and
better performing features can be learned, as demonstrated in \secc{reset}.

This reset may be explained as follows. During alternating
inference and learning stages, the model can overfit a mini-batch of data
by optimizing either the filters or feature maps too much. This 
causes the model to lock up in a state where no new feature map element
can turn on because the reconstruction performance is sufficient to have
only a small error propagating forward to the feature level. Since no new
features turn on after shrinkage, the filters remain fixed as they
continue to get the same gradients. This can happen early in the
learning procedure when the filters are still not optimal and
therefore the learned representation suffers. By resetting the feature
maps, at the next epoch the model has to reevaluate how to reconstruct
the image from scratch, and can therefore turn on the optimal feature
elements and continue to optimize the filters.

\subsection{(Un)pooling Variable Updates} 
\label{sec:w_infer}
Given a model with $l$ layers, we wish to update the (un)pooling
variables $\theta_k$ at each intermediate layer $k$ to optimize the
objective $C_l(v)$.  We assume that the filters $f_1,\ldots,f_l$ and
features $p_l$ are fixed. 

The gradients for the pooling variables $\theta_k$ involve combining,
at layer $k$, the forward propagated error signal with the top down
reconstruction signal. This combined signal then drives the update of
the pooling variables. More formally:
 \BE
\frac{\partial C_l(v)}{\partial U_{\theta_k}} = R_k^T (R_l p_l - v) \cdot
(R_{(l \rightarrow k)} p_l) 
\EE 
where $R_{l \rightarrow k}$ is the top
down reconstruction from layer $l$ feature maps to layer $k$ feature
maps and $R_k^T$ is the error propagation up to $z_k$.

With the chosen Gaussian parameterization of the pooling regions,
the chain rule can be used to compute the gradient for each parameter
$\theta_k = \{\mu_x,\mu_y,\gamma_x,\gamma_y\}$:
\BE
\nabla \theta_k = \frac{\partial C_l(v)}{\partial \theta_k(j)}  = \sum_{i' \in N_j}
\frac{\partial C_l(v)}{\partial U_{\theta_k}(i')} \frac{\partial U_{\theta_k}(i')}{\partial w(i')} \sum_{i \in N_j} \frac{\partial w(i')}{\partial a(i)} \frac{\partial a(i)}{\partial \theta_k(j)} 
\label{eqn:poolgrad}
\EE 
where $j$ is the neighborhood index, 
\BE
\frac{\partial U_{\theta_k}(i')}{\partial w(i')} = \hat{p}_k(i') = (R_{(l \rightarrow k)} p_l)(i')
\EE
 
\BE
\frac{\partial w(i')}{\partial a(i)} = (\sum_{n\in N} a(n))^{-1}[w(i)]
\EE 

\BE
\frac{\partial w(i')}{\partial a(i')} = (\sum_{n \in N} a(n))^{-1}[1-w(i')]
\EE 

\BE
\frac{\partial a(i)}{\partial \mu_x(j)} = \gamma_x(j) (x(i)-\mu_x(j)) a(i)
\EE 

\BE
\frac{\partial a(i)}{\partial \mu_y(j)} = \gamma_y(j) (y(i)-\mu_y(j)) a(i)
\EE 

\BE
\frac{\partial a(i)}{\partial \gamma_x(j)} = -\frac{1}{2} (x(i)-\mu_x(j))^2 a(i)
\EE 

\BE
\frac{\partial a(i)}{\partial \gamma_y(j)} = -\frac{1}{2} (y(i)-\mu_y(j))^2 a(i)
\EE 
where $x(i)$ and $y(i)$ are the coordinates within the pooling
neighborhood $N_j$. 

Once the complete gradient is computed as in \eqn{poolgrad}, we do a gradient step on each pooling variable:
\BE
\theta_k = \theta_k - \lambda_l \beta_{U_k} \nabla \theta_k 
\EE 
using a fixed step size $\beta_{U_k}$. We experimented with a similar step size to \eqn{autostep} for the pooling parameters, however found the estimates to be unstable, likely due to the nonlinear derivatives involved in the Gaussian pooling.

\begin{algorithm}[t!]
\small
\begin{algorithmic}[1]
\REQUIRE Training set $Y$, \# layers $L$, \# epochs $E$, \# ISTA steps $T$
\REQUIRE Regularization coefficients $\lambda_l$, \# feature maps $B_l$ 
\REQUIRE Pooling step sizes $\beta_{U_l}$
\FORC{$l=1:L$}{ \%\% Loop over layers}
\STATE Init.\ features/filters: $p^i_l \sim
0$, $f_l \sim \mathcal{N}(0, \epsilon)$
\STATE Init.\ switches: $\theta^i_l = Fit(R^T_l y_i) \;\; \forall i$
\FORC{epoch $=1:E$}{ \%\% Epoch iteration}
\FORC{$i=1:N$}{ \%\% Loop over images}
\FORC{$t=1:T$}{ \%\% ISTA iteration}
\STATE Reconstruct input: $\hat{v_l}^i = R_l p^i_l$
\STATE Compute reconstruction error: $e=\hat{v_l}^i-v^i$
\STATE Propagate error up to layer $l$: $\nabla p_l = R^T_l e$
\STATE Estimate step size $\beta_{p_l}$ as in \eqn{autostep}
\STATE Take gradient step on p: $p^i_l = p^i_l - \lambda_l \beta_{p_l} \nabla p_l$
\STATE Perform shrink: $p^i_l = \max(|p^i_l|-\beta_{p_l},0) \text{sign}(p^i_l)$ 
\STATE Project to positive: $p^i_l = \max(p^i_l,0)$ 
\FORC{$k=1:l$}{ \%\% Loop over lower layers}
\STATE Take gradient step on $\theta$: $\theta^i_k = \theta^i_k - \lambda_l \beta_{U_k} \nabla \theta_k$
\ENDFOR
\ENDFOR
\ENDFOR
\STATE Update $f_{l}$ by solving \eqn{cg_f} using CG
\STATE Project $f_{l}$ to positive and unit length
\ENDFOR
\ENDFOR 
\STATE Output:  filters $f$, feature maps $p$ and pooling variables $\theta$.
\end{algorithmic}
\small
\caption{Learning with Differentiable Pooling in Deconvolutional Networks}
\small
\label{alg:am} 
\end{algorithm}

\section{Learning}
\label{sec:learning}
\vspace{-2mm}

After inference of the feature maps for the top layer and (un)pooling
variables for all layers is complete, the filters in each layer are
updated. This is done using the gradient with respect to each layer's
filters:
\BE
\frac{\partial C_l}{\partial (f_c^b)_k} = \lambda_l [R_{k-1}^T(R_l p_l - v)]^c \conv [(U_{\theta_k} R_{(l \rightarrow k)} p_l)]_b
\label{eqn:cg_f}
\EE
where the left term is the bottom up error signal propagated up to the feature maps below the given filters, $p_{k-1}$ and the right term is the top down reconstruction to the unpooled feature maps $z_k$. The gradient is therefore the convolution between all combinations of input error maps to the layer (indexed by $c$) and the unpooled feature maps reconstructed from above (indexed by $b$), resulting in updates of each filter plane $f_c^b$, for each layer $k$.

In practice we use batch conjugate gradient updates for learning the filters as
the model is linear in $F_k$ once the feature maps and pooling
parameters are inferred. After 2 steps of conjugate gradients, the
filters are projected to be nonnegative and renormalized to unit
$\ell_2$ length.

\subsection{Joint Inference}
The objective function explicitly constrains the reconstruction from
the top layer features to be close to the input image. From this we
can calculate gradients for each layer's filters and pooling variables
while optimizing the top level features maps. Therefore for each image
we can infer the local shifts and scalings of low level features as
the high level concepts develop.

We have found that pre-training the first layer in one phase of
training and then using the pooling variables and learned layer 1
filters to initialize a second phase of training works best. The
second phase of training optimizes the second layer objective from
which we can update $p_2$, $U_{w_2}$, $U_{w_1}$, $F_2$, and $F_1$
jointly. If care is not taken in this joint update, the first layer
features can trade off representation power with the second layer
filters. This can result in the second layer filters capturing the
details while the first layer filters become dots. To avoid this
problem, after the first phase of training we hold $F_1$ fixed and
optimize the remaining variables jointly. Thus, while the filters are
learned layer-by-layer, inference is always performed jointly across
all layers. This has the nice property
that these low level parts can move and scale as the $U_{w_1}$
variables are optimized while the high level concepts are learned.

\vspace{-3mm}
\section{Initialization of Parameters}
\label{sec:init}
Before training, the filter parameters are initialized to Gaussian distributed random values. After this random initialization, the filters are projected to be non-negative and normalized to unit length before training begins. 

Before inference, either at the start of training or at test time, we initialize the features maps to 0. This creates a reconstruction of 0 in the pixel space, therefore the initial gradient being propagated up the network is $-y$. This is similar to a feedforward network for the first iteration of inference. While forward propagating this signal up the network we can leverage the Gaussian parameterization of the pooling regions to fit these pooling parameters using moment matching. That is, at each layer, we extract the optimal pooling parameter that fit this bottom up signal. This provides a natural initialization to both the pooling variables at each layer and the top level feature activations given the input image and the filter initialization.

\vspace{-2mm}
\section{Experiments}
\vspace{-1mm}
{\bf Evaluation on MNIST}
We choose to evaluate our model on the MNIST handwritten digit classification task. This dataset provides a relatively large number of training instances per class, has many other results to compare to, and allows easy interpretation of how a trained model is decomposing each image.

{\bf Pre-processing:} 
The inputs were the unprocessed MNIST digits at 28x28 resolution. Since no preprocessing was done, the elements remained nonnegative.

{\bf Model architecture:} 
We trained a 2 layer model with 5x5 filters in each layer and 2x2 non-overlapping pooling regions. The first layer contained 16 feature maps and the second layer contained 48 features maps. Each of these 48 feature maps connect randomly to 8 different layer 1 feature maps through the second layer filters. These sizes were chosen comparable to \cite{Zeiler11} while being more amenable to GPU processing. The receptive fields of the second layer features are 14x14 pixels with this configuration, or one quarter the input image size. 

{\bf Classification:} 
One motivation of this paper was to analyze how the classification pipeline of Zeiler \etal \cite{Zeiler11} could be simplified by making the top level features of the network more informative. Therefore, in this paper we simply treat the top level activations inferred for each image as input to a linear SVM \cite{LibLinear}. 

The only post processing done to these high level activations is that overlapping patches are extracted and pooled, analogous to the dense SIFT processing which is shown by many computer vision researchers to improve results \cite{Ylan}. This step provides an expansion in the number of inputs, allowing the linear SVM to operate in a higher dimensional space. For layer 1 classification these patches were 9x9 elements of the layer 1 features maps. For layer 2 they were 6x6 patches, roughly the same ratio to the feature map size as for layer 1. These patches were concatenated as input to the classifier. Throughout the experiments we did not combine features from multiple layers, concatenating only layer 1 patches together for layer 1 classification and only layer 2 features together for layer 2 classification. These final inputs to the classifier were each normalized to unit length.

{\bf Hyperparameters:}
By cross validating on a 50,000 train and 10,000 validation set of
MNIST images, we found that $\lambda_1 = 2$ and $\lambda_2 = 0.5$ gave
optimal classification performance. Each layer was trained with 100
ISTA steps/epoch for 50 epochs (passes through the dataset). After epoch 25, the feature maps were reset to 0 during training. At test time, we found higher $\lambda_1=5$ and $\lambda_2=5$ improved classification, as did optimizing for only 50 ISTA steps of inference.

\begin{figure}[h!]
\begin{center}
\includegraphics[width=3.0in]{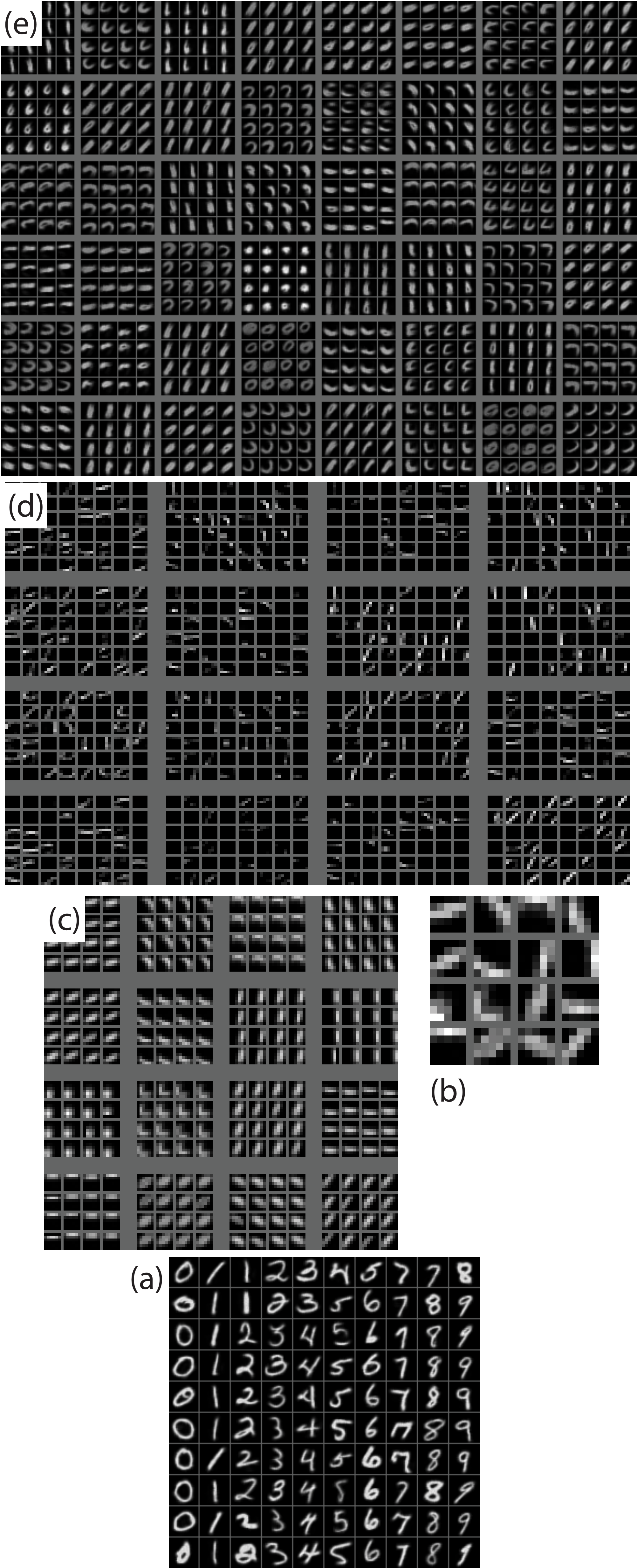}
\end{center}
\vspace*{-0.3cm}
\caption{Visualization of the trained model: (a) reconstructions from layer 2, (b) the 16 layer 1 filter weights, (c) invariance visualization for layer 1 incorporating unpooling and convolutions (see \secc{composite} for details) (d) layer 2 filter weights (shown as 16 groups of filter planes connecting to all 48 layer 2 maps), (e)  layer 2 pixel space invariance visualization of features projected down from samples of the layer 2 feature distribution (see \secc{composite}).}
\label{fig:composite} 
\end{figure}

\subsection{Model visualization}
\label{sec:composite} 
\vspace{-2mm}
By visualizing the filters and features maps of the model, we can
easily understand what it has learned. In \fig{composite} (a) we
demonstrate sharp reconstructions of the input images from the second
layer features maps. In \fig{composite} (b) we display the raw filter
coefficients for layer 1 which have learned small pieces of strokes. By incorporating the pooling parameters into the layer, these filters are robust to small changes in the input. 

Visualizing these invariances of a model can be helpful in
understanding the inputs the model is sensitive to. Searching through
the dataset of inferred feature map activations and selecting the
maximum element per feature map to project downward into the pixel
space as in \cite{Zeiler11} is one way of visualizing these
invariances. However, these selected elements are only exemplars of
inputs that most strongly activated that feature. In
\fig{composite}(c) we show a more representative selection of
invariances by instead selecting a feature activation to be
projected down based on sampling from the distribution of
activations for that feature inf the dataset. This gives a less biased
view of what activates that feature than selecting the largest few activations from the dataset. Once a sample is selected for a given feature map, the pooling variables corresponding to the image from which the activation was selected are used in the unpooling stages to do the top down visualization. 

Examining the 16 sample visualizations for each feature in \fig{composite}(c) shows the scale and shifts that the Gaussian pooling provides to these relatively simple first layer filters. We can continue to analyze the model by viewing the layer 2 filters planes in \fig{composite}(d). Each of the 48 second layer features has 16 filter planes (shown in separate groups), one connecting to each of the layer 1 feature maps. While the second layer filters are difficult to understand directly, we can visualize the learned representation of the second layer by projecting down all the way to the pixel space through layer 1. \fig{composite}(e) shows for each of the 48 feature maps a 4x4 grid of pixel space projections obtained by sampling 16 activations from the distribution of activations of each layer 2 feature and projecting down via alternating convolution and unpooling with the corresponding pooling variables separately for each activation.

While analyzing the features in pixel space is informative, we have
also found it is useful to view the features as decompositions of an
input image to know how the model is representing the data. One
possible method of displaying the decomposition is by coloring each
pixel of the reconstruction according to which feature it came
from. Each feature is assigned a hue (in no particular order) and the
associated reconstruction produced then defines the saturation of that
color. The resulting image therefore depicts the high level feature
assignments. Pixels with brownish colors indicate a summation of
several colors (features) together. Note that the input images themselves are
grayscale -- the colors are just for visualization purposes.

\label{sec:vis}
\begin{figure}[t!]
\begin{center}
\includegraphics[width=2.9in]{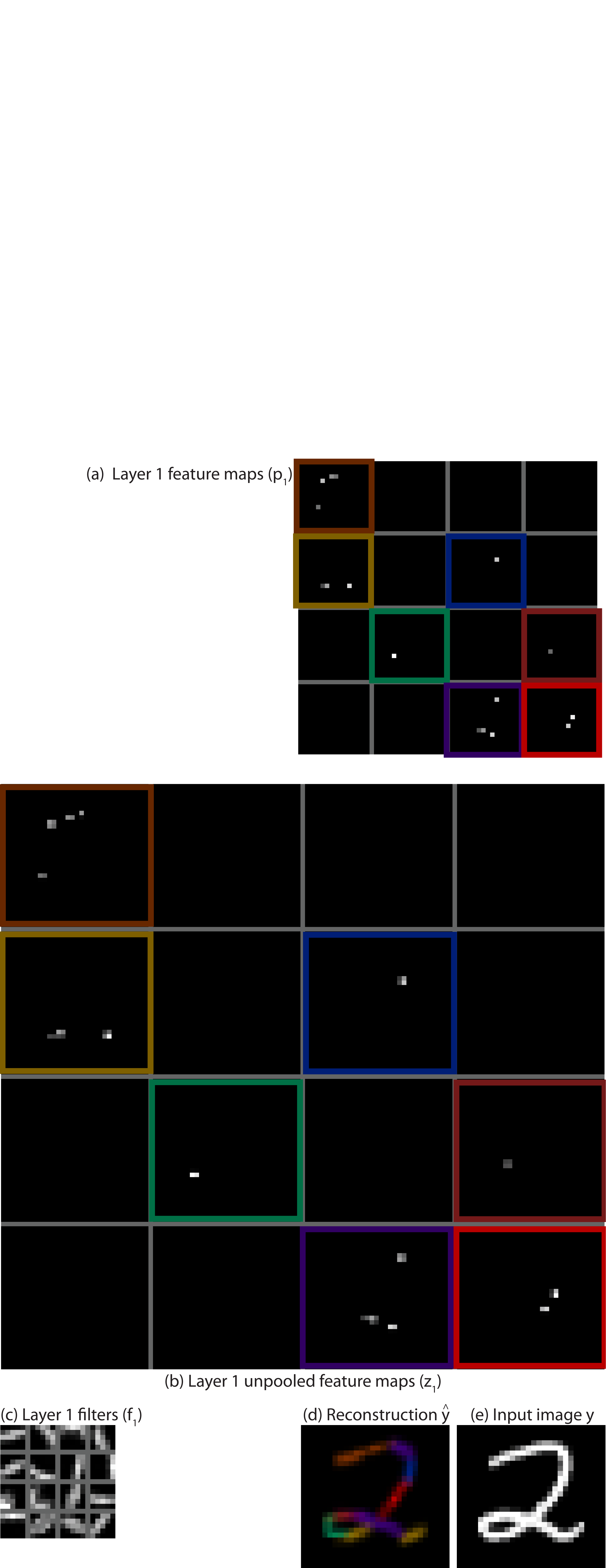}
\end{center}
\vspace*{-0.3cm}
\caption{One layer decomposition of a digit into parts. From the top down the layer 1 feature maps (a) are unpooled into (b) and convolved with (e) to produce the reconstruction (f). The colors in the reconstruction simply represent which feature the reconstructed pixel came from.}
\label{fig:decomp_l1} 
\vspace*{0.3cm}
\end{figure}

\begin{figure}[t!]
\begin{center}
\includegraphics[width=2.9in]{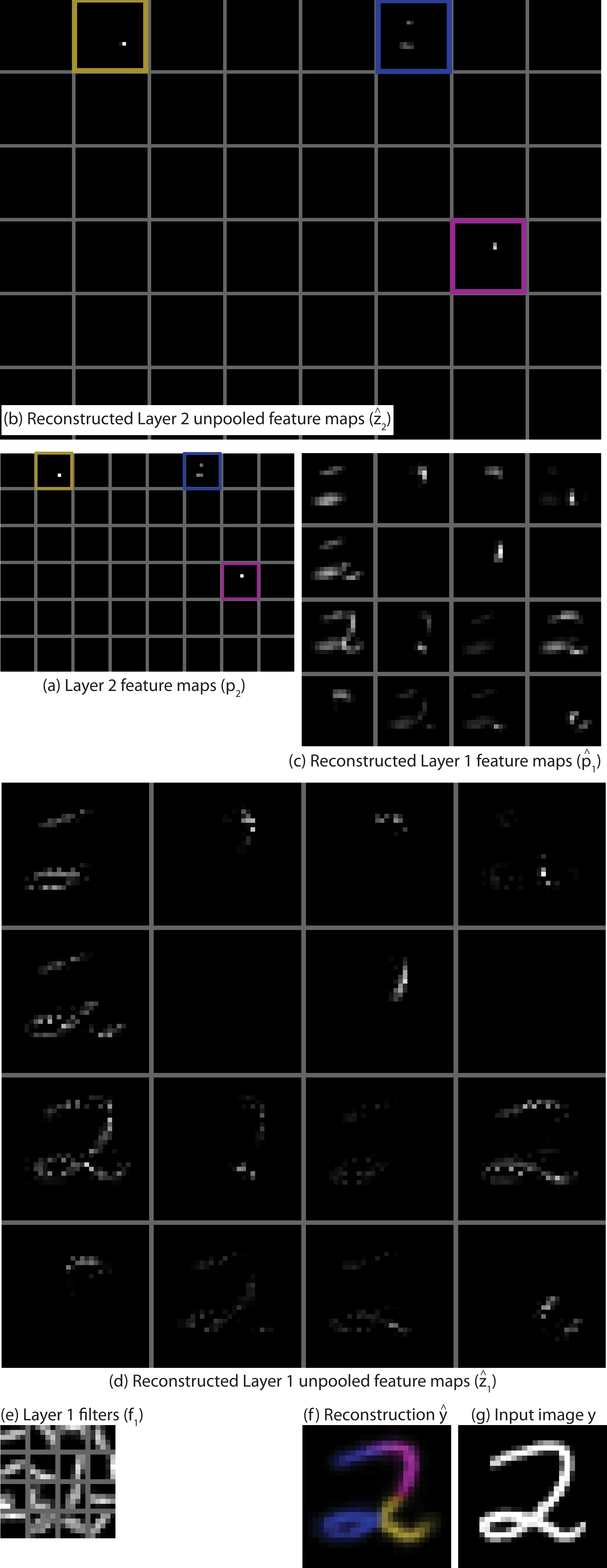}
\end{center}
\vspace*{-0.3cm}
\caption{Two layer decomposition of a digit into parts. From the top down the layer 2 feature maps (a) are unpooled into (b) and convolved with the layer 2 filters to produce the reconstruction of layer 1 feature maps (c). These are unpooled into (c) and convolved with (e) to produce (f), colored according to the layer 2 feature that the reconstructed pixel it was reconstructed from.}
\label{fig:decomp_l2} 
\end{figure}

In \fig{decomp_l1}(d) we show such a reconstruction from layer 1 for the original image in (e). To understand the model we also show the layer 1 feature map activations in (a) with their corresponding color assignment around them. Notice the sparse distribution of activations can reconstruct the entire image by utilizing the Gaussian pooling and layer 1 filters in (c). \fig{decomp_l1}(b) shows the result of this unpooling operation on the feature maps. Notice in the orange and purple boxes the elongated lines in the unpooled maps, made possible by a low precision in one dimension. 

\fig{decomp_l2} takes this analysis one step further by using the
second layer of the model. Starting from 3 features in the layer 2
feature maps as shown in (a), they are unpooled (as shown in (b)) and
then convolved with the second layer filters to reconstruct many elements down on to the first layer features maps (c). These are further unpooled to (d) where again you can see the benefits of the Gaussian pooling smoothly transitioning between non-overlapping pooling regions. These are finally convolved with first layer filters (e) to give the decomposition shown in (f). Notice how long range structures are grouped into common features in the higher layer compared to the layer 1 decomposition of \fig{decomp_l1}.

\subsection{Max Pooling vs Gaussian Pooling}
\label{sec:max}

\begin{figure}[h!]
\begin{center}
\end{center}
\begin{center}
\includegraphics[width=3.1in]{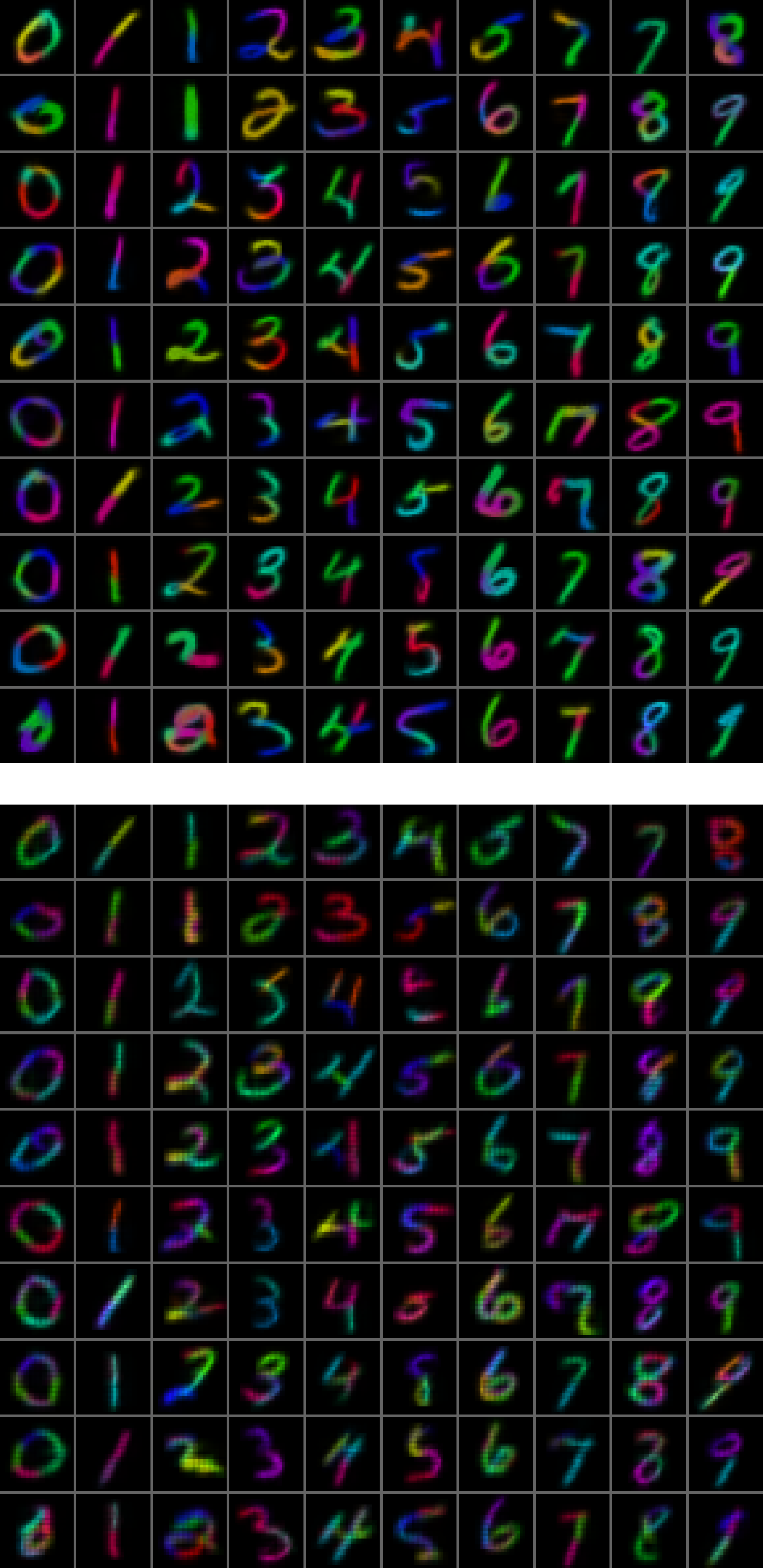}
\end{center}
\vspace*{-0.3cm}
\caption{Feature decomposition comparison between Gaussian pooling
  (top) and max pooling (bottom). Each reconstructed pixel's color
  corresponds to the layer 2 feature map it was reconstructed
  from. Note the reuse of similar strokes in digits of a different
  class. Aliasing artifacts are present in the reconstructions using max
  pooling -- see \secc{max}. }
\label{fig:maxgauss} 
\end{figure}

The discrete locations that max pooling allows within a region are a limiting factor in the reconstruction quality of the model. \fig{maxgauss} (bottom) shows a significant aliasing effect is present in the visualizations of the model when Max pooling is used. With the complex interactions between positive and negative elements removed, the model is not able to form smooth transitions between non overlapping pooling regions even though the filters used in the succeeding convolution sublayer have overlap between regions. Using the Gaussian pooling, the model can infer the desired precisions and means in order to optimize the reconstruction quality from high layers of the model. 

\begin{figure}[t!]
\begin{center}
\includegraphics[width=3.1in]{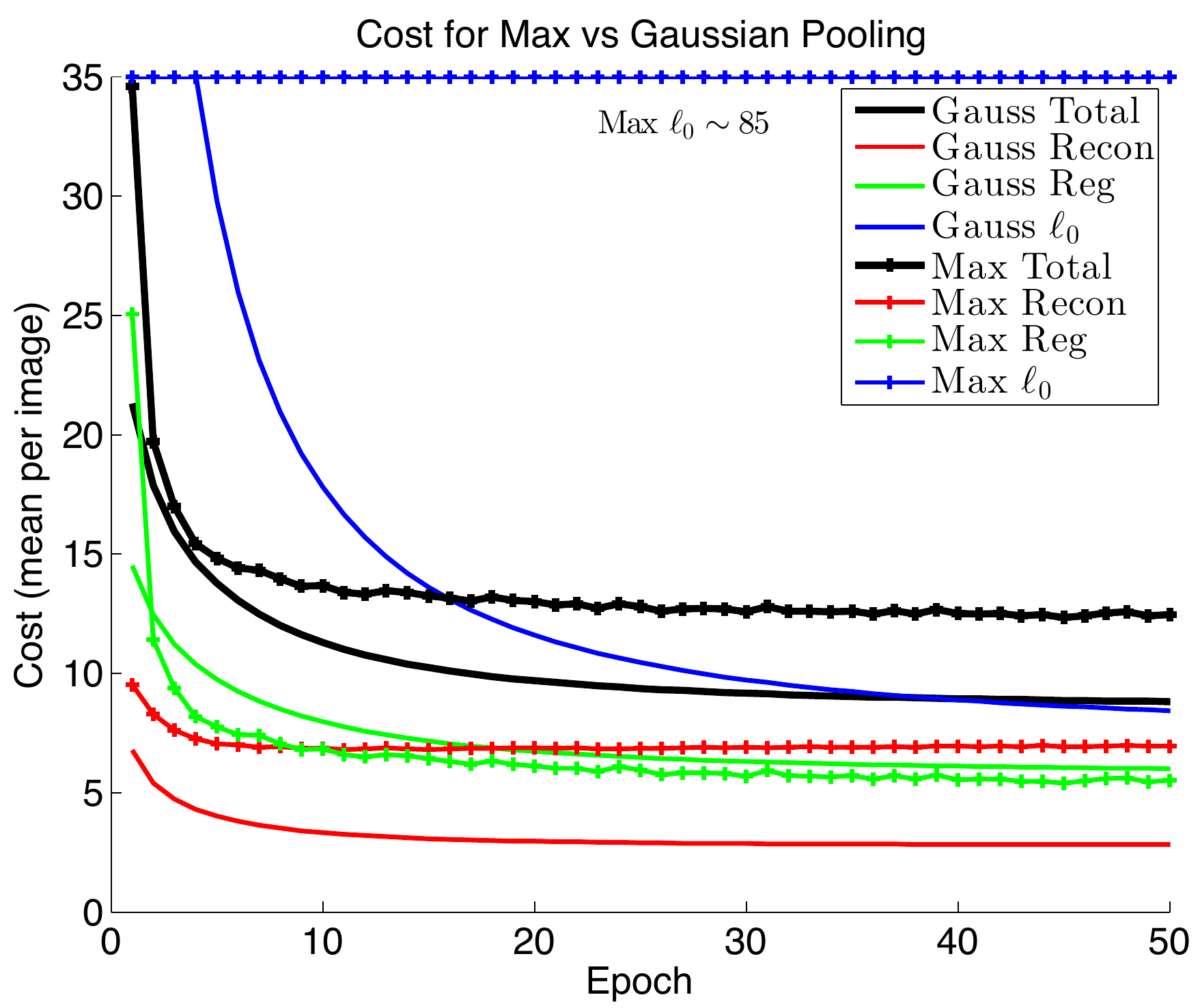}
\end{center}
\vspace*{-0.3cm}
\caption{Breakdown of cost function into reconstruction and
  regularization terms for Max and Gaussian pooling for 2 layer models. Gaussian pooling
  gives consistently lower cost than max pooling. Furthermore, the
  $\ell_0$ sparsity (shown in blue), is significantly lower for the Gaussian
  pooling, although not explicitly part of the cost.}
\label{fig:costplot}
\vspace*{-4mm}
\end{figure}

This fine tuning of reconstruction allows for improvements without significantly varying the features activations (ie. maintains or decreases the sparsity while adjusting the pooling parameters). This is confirmed in \fig{costplot} where we break down the cost function into the reconstruction and regularization terms. In this figure we also display the $\ell_0$ sparsity of each model as this can directly be used for comparison. 

The Gaussian pooling significantly outperforms Max pooling in terms of optimizing the objective. By not being able to adjust the pooling variables to optimize the overall cost, Max pooling plateaus despite running for many epochs. Additionally it has a much higher $\ell_0$ cost throughout training. In contrast, the $\ell_0$ cost with Gaussian pooling decreases smoothly throughout training because the model can fine tune the pooling parameters to explain much more with each feature activation. This property is shown in \tab{maxgauss} to significantly improve classification performance compared to Max pooling when stacking.
 
\begin{table}[h!]
\small
\vspace*{-0mm}
\begin{center}
\begin{tabular}{|l|c|c|}
  \hline
    & Layer 1 & Layer 2\\
  \hline 
   Max Pooling & $1.30\%$ & $1.25\%$ \\
   Gaussian Pooling & $1.38\%$  & $0.84\%$\\
   \hline 
\end{tabular}
\vspace*{1mm}
\caption{MNIST error rate of Max pooling versus Gaussian pooling for 1 and 2 layer models. Note the performance improvement when stacking layer with Gaussian pooling.}
\label{tab:maxgauss}
\vspace*{-5mm}
\end{center}
\end{table}

\vspace{4mm}
\subsection{Joint Inference}
\label{sec:joint}
One of the main criticisms of sparse coding methods is that inference must be conducted even at test time due to the lack of a feedforward connection to encode the features. In our approach we discovered two fundamental techniques that mitigate this drawback. 

The first is that running a joint inference procedure over both layers
of our network improves the classification performance compared to
running each layer separately. Instead of inferring the feature maps
and pooling variables for the first layer and then using these pooling
variables to initialize the second layer inference (2 phases), we can
directly run inference with a two layer model. The differentiable
pooling allows us to infer the pooling variables of both layers in
addition to the layer 2 feature values simultaneously in 1 phase. At the first iteration of inference we leverage the ability to fit the Gaussian pooling parameters in a feed forward way as mentioned in \secc{init}. This halves the number of inference iterations needed by not requiring any first layer inference prior to inferring the second layer.

To examine this first discovery in depth we considered several combinations of how to joint train and then run inference at test time with this model. During training we have found both qualitatively in terms of feature diversity and quantitatively in terms of classification performance that training in separate phases, one for each layer of the model, works better than jointly training both layers from scratch. In the second phase of training, when optimizing for reconstruction from the second layer feature maps, the first layer pooling variables and filters can either be updated or held fixed. Each row of \tab{1phase2phase} examines each combination of these updates during training. We can see that the optimal training scheme was with fixed first layer filters but pooling updates on both layers. This made the system more stable while still allowing these first layer filters to move and scale as needed by updating the first layer pooling variables.

In all cases we see a significant reduction in error rates when doing inference in 1 phase. The middle column of the table shows this 1 phase inference, but without optimizing the first layer pooling parameters $U_1$ whereas the last column does optimize $U_1$. We see an improvement in updating $U_1$ for all but the last row which was trained without $U_1$ and so is used to that type of inference. This improvement with joint inference of $U_1, U_2,$ and $p_2$ is a key finding which is only possible with differentiable pooling.

\begin{table}[h!]
\small
\vspace*{-0mm}
\begin{center}
\begin{tabular}{|l|c|c|c|}
  \hline
    Training & Infer 2 & Infer 1 & Infer 1 \\
     & phases & phase (no $U_1$) & phase \\
  \hline 
   Updating $F_1$ $U_{w_1}$ & $1.79\%$  & $1.63\%$ & $1.40\%$ \\
   Updating $F_1$ & $1.71\%$  & $1.21\%$ & $1.10\%$  \\
   Updating $U_{w_1}$ & $1.39\%$  & $1.04\%$ & $0.84\%$ \\
   No Layer 1 Updates & $1.46\%$ & $0.99\%$ & $1.03\%$ \\
   \hline 
\end{tabular}
\vspace*{1mm}
\caption{Comparison of joint training techniques. Each row is a
  trained two layer model that updates select variables in layer 1
  during training (in addition to $F_2$ and $U_{w_2}$). The three
  columns use these models but run inference at test time in 2 phases, 1 phase without updating $U_1$, and 1 phase with all updates respectively.}
\label{tab:1phase2phase}
\vspace*{-5mm}
\end{center}
\end{table}

\begin{figure}[h|]
\vspace{-3mm}
\begin{center}
\includegraphics[width=3.1in]{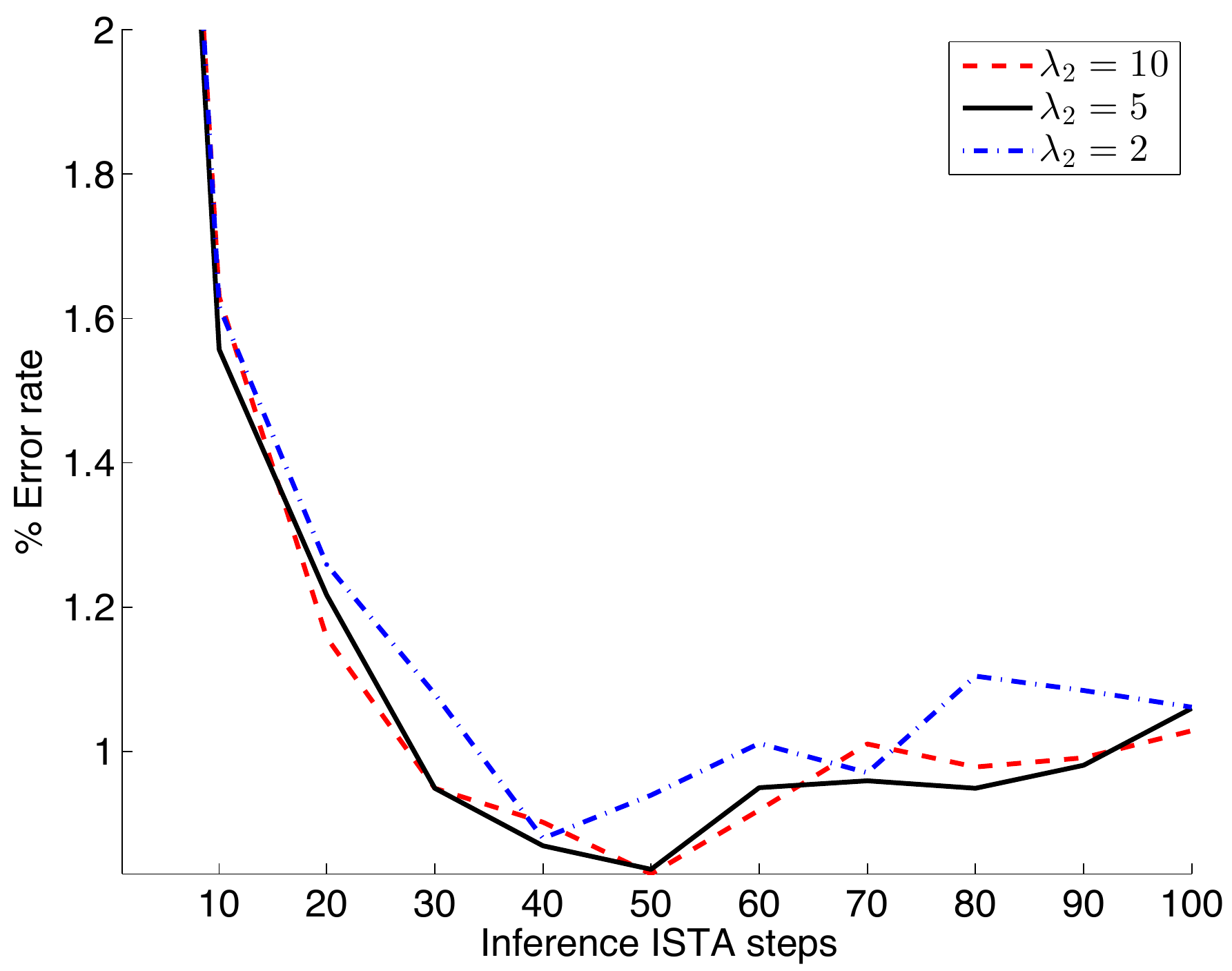}
\end{center}
\vspace*{-0.3cm}
\caption{Comparison of classification errors versus number of ISTA steps used during inference.}
\label{fig:ista} 
\vspace{-5mm}
\end{figure}

The second discovery that reduces evaluation time is that running the
same number of ISTA iterations as was done during training does not
give optimal classification performance, possibly due to
over-sparsification of the features. Similarly running with too few iterations also reduces performance. \fig{ista} shows a plot comparing the number of ISTA iterations to the classification performance with an optimum at 50 ISTA steps, half the number used during training.

\vspace{5mm}
\subsection{Effects of Non-Negativity}
\label{sec:pos}
With negative elements present in the system, many possible solutions can be found during optimization. This happens because subtractions allow the removal of portions of high level features. This has the effect of making them less discriminative because the model can change parameters in-between the high level feature activations and the input image in order to reconstruct better while assigning less meaning to the feature activations themselves.

To show this is not an artifact of the Gaussian pooling being more suited to nonnegative systems (due to the summation over the pooling region possibly leading to cancellations if negatives are present), we include comparison in \tab{posneg} to Max pooling. In both cases, enforcing positivity via projected gradient descent improves the discriminative information preserved in the features. 

\begin{table}[h!]
\small
\vspace*{-0mm}
\begin{center}
\begin{tabular}{|l|c|c|}
  \hline
    & Positive/Negative & Non-negative \\
  \hline 
   Max Pooling & $2.04\%$ & $1.25\%$ \\
   Gaussian Pooling & $2.32\%$  & $0.84\%$ \\
   \hline 
\end{tabular}
\vspace*{1mm}
\caption{MNIST error rate for Max and Gaussian models trained with and without the non-negativity constraint.}
\label{tab:posneg}
\vspace*{-5mm}
\end{center}
\end{table}

\vspace{1mm}
\subsection{Effects of Feature Reset} 
\label{sec:reset}
\begin{figure}[h!]
\vspace*{-0.3cm}
\begin{center}
\includegraphics[width=3.1in]{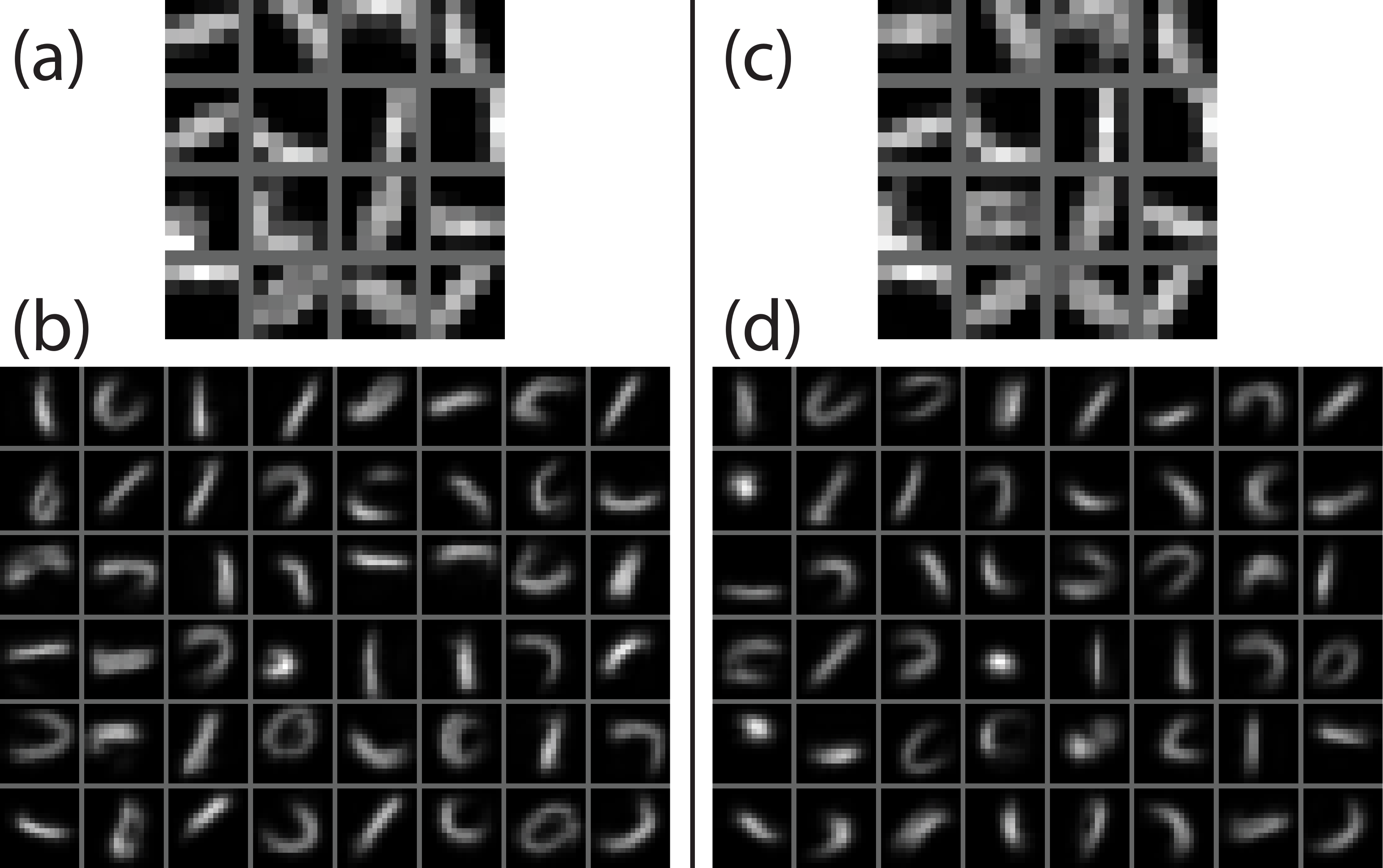}
\end{center}
\vspace*{-0.3cm}
\caption{Qualitative difference in first layer filters with (left) and without (right) resetting of the feature maps.}
\label{fig:reset} 
\end{figure}

When training the model on MNIST, some less than optimal filters are learned when not resetting the feature maps. For example, in \fig{reset} (c) many of these layer 1 filters are block-like such as the 3rd row, 2nd column. However this same feature in (a) improves if the feature maps are reset to 0 once half way through training. This single reset is enough to encourage the filters to specialize and improve. 
Similarly, the layer 2 pixel visualizations in (b) have much more variation due to the reset compared to (d) which did not have the reset. In particular, notice many blob-like features learned in (d) without reset such as the 2nd and 5th rows of the 1st column that improve in (b). These larger, more varied features learned with the reset help improve classification performance as shown in \tab{reset}.

\begin{table}[h!]
\small
\vspace*{-3mm}
\begin{center}
\begin{tabular}{|c|c|}
  \hline
 Trained with No Reset & Trained with Reset \\
  \hline 
   $1.00\%$ & $0.84\%$ \\
   \hline 
\end{tabular}
\vspace*{1mm}
\caption{MNIST error rates for 2 layer models trained with and without resetting the feature maps.}
\label{tab:reset}
\vspace*{-5mm}
\end{center}
\end{table}

\vspace{2mm}
\subsection{Effects of Hyper-Laplacian Sparsity} 
\label{sec:sparsity}
\begin{figure}[h!]
\vspace*{-3mm}
\begin{center}
\includegraphics[width=3.1in]{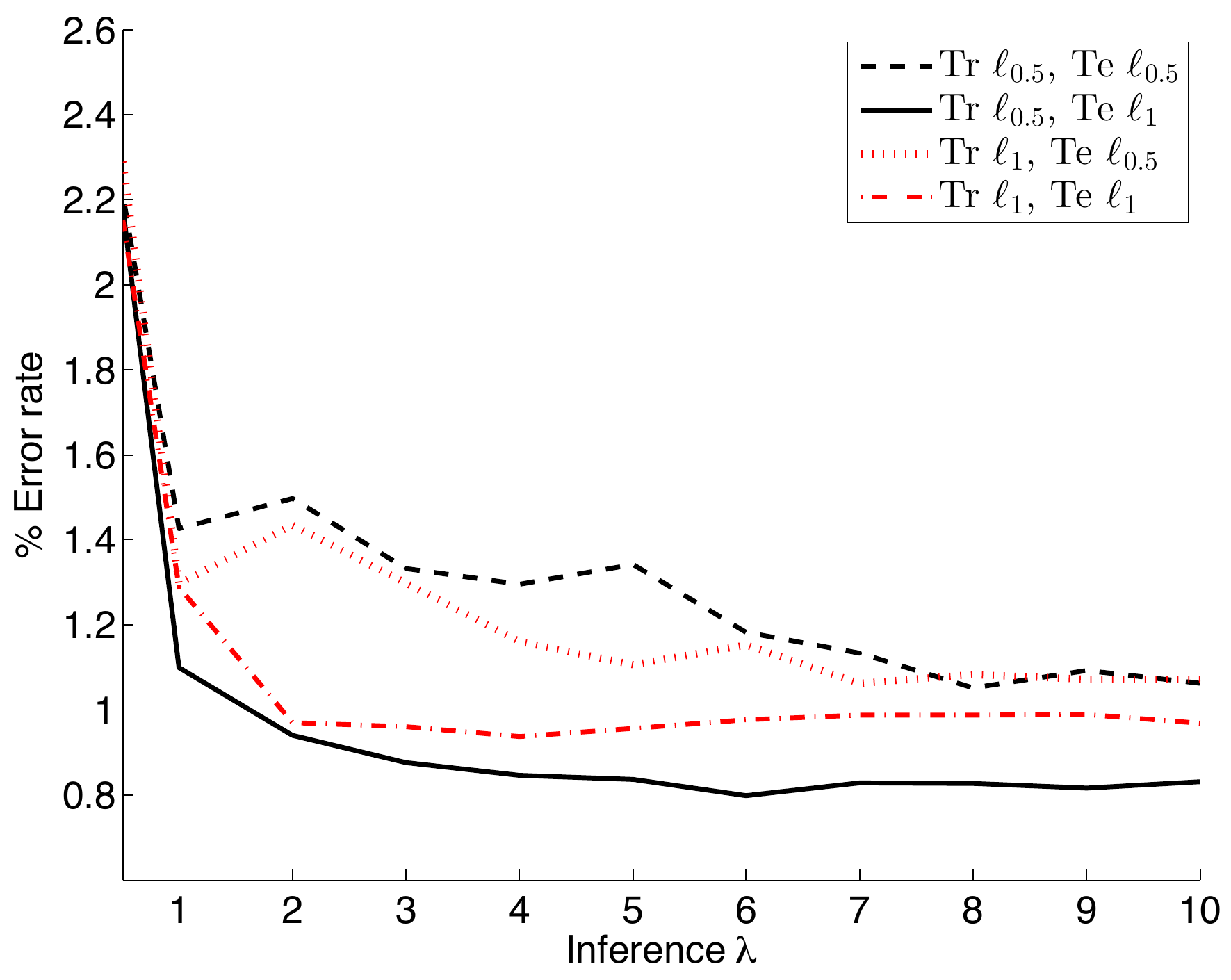}
\end{center}
\vspace*{-0.3cm}
\caption{Error rates for $\ell_{1}$ and $\ell_{0.5}$ priors used in training and inference.}
\label{fig:priors} 
\vspace*{-3mm}
\end{figure}

It has previously been shown that sparsity encourages learning of distinctive features, however it is not necessarily useful for classification \cite{Rigamonti11} \cite{Zeiler11}. We analyze this in the context of hyper-laplacian sparsity applied to both training and inference. In this comparison we trained two models, one with a $\ell_{1}$ prior on the feature maps and the other with a $\ell_{0.5}$ prior. Once trained, we took each model and ran inference with both $\ell_{1}$ and $\ell_{0.5}$ priors. For reference the $\ell_0$ sparsity for the training runs was 4.2 for the $\ell_{0.5}$ regularized training and 20.2 for the $\ell_{1}$ regularized training with the same $\lambda_2 = 0.5$ setting. Since the amount of sparsity can also be controlled during inference by the $\lambda_2$ parameter, we plot in \fig{priors} the classification performance for various $\lambda_2$ settings in these four model combinations.

Interestingly, utilizing the added sparsity during training enforced by the $\ell_{0.5}$ while using the more relaxed $\ell_{1}$ prior for inference is the optimal combination for all $\lambda$ settings. This suggests sparsity is useful during training to learn meaningful features, but is not as useful for inference at test time.

\subsection{Comparison to Other Methods}
\vspace{-2mm}
\label{sec:compare}
We chose the MNIST dataset for it's large number of results to compare to. Of these, deep learning methods typically fall into one of two categories, 1) those that are completely unsupervised and have a simple classifier on top, or 2) those that are fine-tune discriminatively with labels. Our method falls into the first category as it is completely unsupervised during training, and only the linear SVM applied on top has access to the label information of the training set. We do not back propagate this information through the network, but this would be an interesting future direction to pursue. \tab{other} shows our method is competitive with other deep generative models, even surpassing several which use discriminative fine tuning.

\begin{table}[t]
\small
\vspace*{-0mm}
\begin{center}
\begin{tabular}{|l|c|c|}
  \hline
    & Pre-training & Fine-tuning \\
  \hline 
   Our Method & $0.84\%$  & -- \\
   CDBN (1+2 layers) \cite{Lee2009} & $0.82\%$ & -- \\
   DBN (3 layers) \cite{Hinton2006a} \cite{Hinton2006b} & $2.5\%$ & $1.18\%$ \\
   DBM (2 layers) \cite{SalHinton07} & -- & $0.95\%$ \\
   \hline 
\end{tabular}
\vspace*{1mm}
\caption{MNIST errors rates for related generative models.}
\label{tab:other}
\vspace*{-5mm}
\end{center}
\end{table}

\section{Discussion}
In this work we introduced the concept of differentiable pooling for deep learning methods. Also, we demonstrated that joint training the model improves performance, positivity encourages the model to learn better representations, and that there is an optimal amount of sparsity to be used during training and inference. Finally, we introduced a simple resetting scheme to avoid local minimum and learn better features. 
We believe many of the approaches and findings in this work are applicable not only to Deconvolutional Networks but also to sparse coding and other deep learning methods in general.


{\small
\bibliographystyle{ieee}
\bibliography{deconv_arxive}
}

\end{document}